\documentclass[journal]{IEEEtran}

\usepackage{times}
\usepackage{epsfig}
\usepackage{graphicx}
\usepackage{amsmath,mathtools}
\usepackage{amssymb}
\usepackage{booktabs}
\usepackage{multirow}
\usepackage{color, colortbl}
\usepackage{arydshln} %dashed line
\usepackage{subfigure}
\DeclareMathOperator*{\argmax}{arg~max}
\DeclareMathOperator*{\argmin}{arg~min}
\usepackage[table]{xcolor}
\usepackage{pifont}
\definecolor{lightgray}{gray}{0.9}
\usepackage{cite} %with IEEEtran.bst, this will collapse the range of refs

\usepackage{soul} %for \hl
\soulregister\cite7
\soulregister\ref7
\soulregister\pageref7
\makeatletter %% <- make @ usable in command sequences
\newcount\SOUL@minus
\makeatother  %% <- revert @

\graphicspath{{IMAGES/}}

\usepackage[pagebackref=true,breaklinks=true,letterpaper=true,colorlinks=true,bookmarks=false]{hyperref}
\usepackage[bookmarks=false]{hyperref}

\begin{document}
\title{BIGPrior: Towards Decoupling Learned Prior Hallucination and Data Fidelity in Image Restoration}

\author{Majed El Helou,~\IEEEmembership{Member,~IEEE,}
Sabine S{\"u}sstrunk,~\IEEEmembership{Fellow,~IEEE.}

\thanks{Both authors are with the School of Computer and Communication Sciences, EPFL, Lausanne, Switzerland. 

{Contact author's email address: \tt\small majed.elhelou@epfl.ch }}
}

\markboth{IEEE TRANSACTIONS ON IMAGE PROCESSING}%
{Shell \MakeLowercase{\textit{et al.}}: Bare Demo of IEEEtran.cls for IEEE Journals}

\newcommand{\subfigBIGP}[2]{ \subfigure[#1]{ \includegraphics[width=0.135\linewidth, trim={0 0 0 0}, clip]{#2} } }
\newcommand{\subfigBIGPphi}[2]{ \subfigure[#1]{ \includegraphics[width=0.135\linewidth, trim={43 43 43 43}, clip]{#2} } }
\newcommand{\BIGPvert}{\vspace{-0pt}}

\newcommand{\medits}[1]{{#1}}
\newcommand{\meditstwo}[1]{{{#1}}} %\hl{}
\definecolor{Gray}{gray}{0.9}

\maketitle

\begin{abstract}
\medits{Classic image-restoration algorithms use a variety of priors, either implicitly or explicitly. Their priors are hand-designed and their corresponding weights are heuristically assigned. Hence, deep learning methods often produce superior image restoration quality. Deep networks are, however, capable of inducing strong and hardly predictable hallucinations. Networks implicitly learn to be jointly faithful to the observed data while learning an image prior; and the separation of original data and hallucinated data downstream is then not possible. This limits their wide-spread adoption in image restoration. Furthermore, it is often the hallucinated part that is victim to degradation-model overfitting.

We present an approach with decoupled network-prior based hallucination and data fidelity terms. We refer to our framework as the Bayesian Integration of a Generative Prior (BIGPrior). Our method is rooted in a Bayesian framework and tightly connected to classic restoration methods. In fact, it can be viewed as a generalization of a large family of classic restoration algorithms. We use network inversion to extract image prior information from a generative network. We show that, on image colorization, inpainting and denoising, our framework consistently improves the inversion results. Our method, though partly reliant on the quality of the generative network inversion, is competitive with state-of-the-art supervised and task-specific restoration methods. It also provides an additional metric that sets forth the degree of prior reliance per pixel relative to data fidelity.}
%Indeed, the per pixel contributions of the decoupled data fidelity and prior terms are readily available in our proposed framework.
\end{abstract}

% Note that keywords are not normally used for peerreview papers.

\let\thefootnote\relax\footnotetext{Our code and models are publicly available at\\ \url{https://github.com/majedelhelou/BIGPrior}}

\IEEEpeerreviewmaketitle

\section{Introduction}
%What is image restoration, and presentation of the key terms
\IEEEPARstart{I}{mage} restoration recovers original images from degraded observations. It is based on two fundamental aspects, specifically, the relation to the observed data and the additional assumptions or image statistics that can be considered for the restoration. The relation to the observed data is referred to as \textit{data fidelity}. The remaining information, brought in by the restoration method based on prior assumptions, is referred to as \textit{prior hallucination}. It is termed hallucination because the added information is derived from a general model or assumption and might not faithfully match the sample image.

% What is the general problem, why is it important
The data fidelity and prior terms emerge theoretically in the Maximum A Posteriori (MAP) formulation, but can also be implicitly induced by the restoration algorithms. For instance, non-local means~\cite{buades2005non} and \medits{Block-Matching and 3D filtering (BM3D)}~\cite{BM3D} utilize the prior assumption that there exists different similar patches within an image. Diffusion~\cite{perona1990scale} methods build on local smoothness assumptions. Data fidelity is typically enforced through the squared norm~\cite{krishnan2009fast} that is equivalent to a MAP-based Gaussian noise model.

% The advantage of explicit priors, and shortcomings
Classic image-restoration algorithms often rely on optimizations over explicit priors. An advantage of explicitly defined priors is the ability to easily control the relative relation between the weight of the data fidelity term and the weight ($\beta$) of the prior term. The general approach consists of an optimization
\begin{equation}
	\argmin_x \psi_d(f'(x),y) + \beta \cdot \psi_p(f''(x)),
\end{equation}
where $y$ is the observation, $f'$ and $f''$ are various manipulation functions \medits{to match the degradation model (for $f'$) and for instance to extract specific components such as frequency bands (for $f''$)}, $\psi_d$ enforces the data fidelity, and $\psi_p$ enforces the prior information. The optimal point is the estimate of the original image $x$.
By making the prior term explicit, it is possible to have control over its contribution hence often better intuition and understanding of the reliability of the final restoration result. 
However, we note two shortcomings of these methods and we expand upon them in the following: (1) $\beta$ is not adapted based on the confidence in the fitness of the prior, and (2) the priors are hand-designed heuristics.

% Discussion of the shortcomings of explicit priors
(1) The parameter $\beta$ should be inversely related to the quality of the observed degraded signal, but it should also be directly related to how well the assumed prior corresponds to the input image distribution or statistics. Although some methods, discussed in the section on related work, adjust their priors to the input data, they do not control $\beta$ based on the confidence in the fitness of the prior to the current sample. (2) Recent methods with implicit data-learned priors, notably relying on deep \medits{Convolutional Neural Networks (CNNs)}, outperform the classic methods with hand-designed priors on various image restoration tasks. This is due to the rich prior learned by discriminative networks or generative networks that, with adversarial training, can even learn image distributions to synthesize new realistic photos\medits{~\cite{karras2018progressive,karras2019style,karras2020analyzing}}. It is worth noting, however, that domain-specific prior information can still be explicitly enforced to improve the performance of the networks~\cite{el2020blind,sindagi2020prior}.

% Discussion of the DL shortcomings
One shortcoming of the deep learning methods is the loss of interpretability and control between data fidelity and prior-based hallucination. Given an image restored by a network, it is not possible to know how faithful it is to the observed signal versus how much prior-based hallucination was integrated in the image. \medits{We call prior-based hallucination the information coming from the learned prior rather than being derived from the observed data. These prior-based hallucinations are not always reliable and can be prone to overfitting~\cite{elhelou2020stochastic}.} Hence, it is important to have a grasp of the prior hallucination taking place in the restoration process.

To obtain decoupled prior-based hallucination and data fidelity terms, we propose a novel framework that we call the \medits{{B}ayesian {I}ntegration of a {G}enerative {Prior} (BIGPrior framework)}. We replace the implicit data prior learned in feed-forward restoration networks with an explicit generative-network prior. This prior is then integrated following a MAP setting, where the data fidelity and prior terms are combined with a fusion weight that is adaptive to both. The BIGPrior framework is a generalization of a large family of classic restoration methods where the prior and its contribution weight are both learned, and the weight can adapt to both the signal quality \textit{and} the fitness of the prior to the observed data.

Our framework structurally provides a reliable metric for per-pixel data fidelity in the final output to answer the question, ``How much hallucination is there - at worst - in the output?". We present and analyze this metric by using blind denoising experiments. We also apply our method to various image restoration tasks and show consistent improvements, notably over the direct use of the generative prior, while additionally providing our faithfulness, a.k.a. data fidelity, metric.

\begin{figure*}[t]
	\centering
	\includegraphics[width=\linewidth,trim={82 30 15 20},clip]{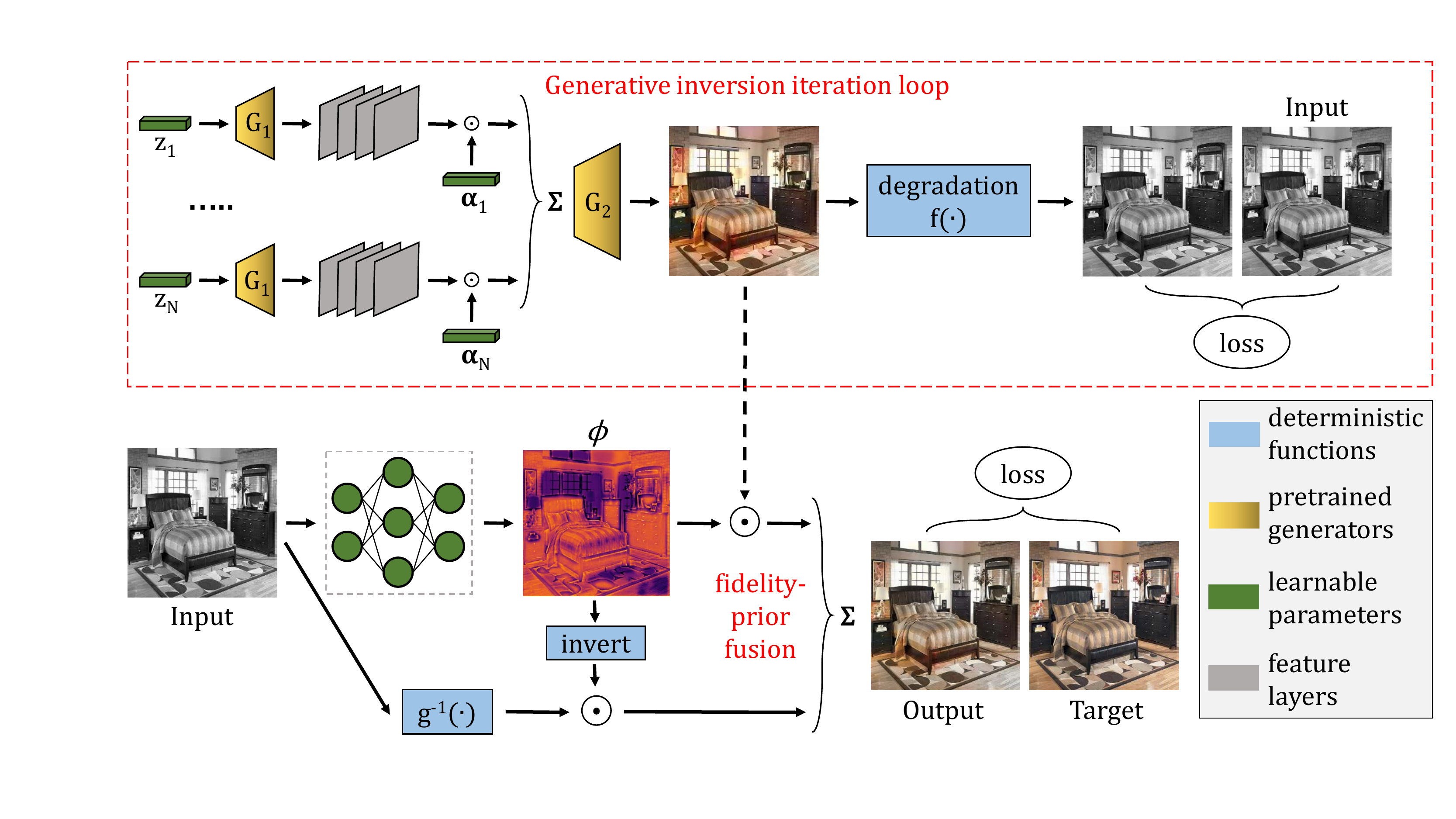}
	\caption[BIGPrior pipeline overview]{Weights that are optimized are shown in green, and the sub-networks of the pre-trained generative network are shown in yellow. The generative network inversion process is optimized over a fixed set of iterations, which regularizes the output~\cite{ulyanov2018deep}. The final output is obtained through the fusion of the prior-based hallucination and the signal information, based on our $\phi$ map estimation.}
	\label{fig:pipeline}
\end{figure*}

\section{Related Work}

\subsection{Classic Image Restoration}
A variety of classic restoration methods, such as Non-Local Means (NLM)~\cite{buades2005non}, BM3D~\cite{BM3D}, their variants~\cite{dabov2009bm3d,lebrun2013nonlocal} or combinations with sparse coding~\cite{dong2014compressive,mairal2009non}, and diffusion-based methods~\cite{perona1990scale,chen2015learning}, make use of various prior assumptions on self-similarity or frequency-content distribution. 
Other algorithms formulate the prior explicitly. For instance, dictionary-based methods~\cite{rubinstein2012analysis} that assume images can be well represented by a fixed set of elements, which we discuss in the next section. Other examples are shrinkage methods~\cite{donoho1995noising,xu2007iterative}. They can be directly connected with the family of MAP estimators, by deriving from the foundational work of Bayes and Laplace~\cite{laplace1998pierre}. Considering an example with a hyper-Laplacian prior on image gradients, originally used in the context of deblurring in~\cite{krishnan2009fast}, optimizing the MAP negative log-likelihood 
\begin{equation}
	\argmin_x -\log(P_{Y|X}(y|x)P_X(x)),
\end{equation}
\medits{where $P_{Y|X}(\cdot,\cdot)$ is the conditional probability distribution of the random variable $Y$ given $X$ and $P_X(\cdot)$ is the probability distribution of $X$}, yields the estimator
\begin{equation}
	\hat{x} = \argmin_x||x-y||_2^2 + \beta \cdot \sum_{j=1}^{J}|x\otimes f_j|^\gamma,
\end{equation}
where $y$ is the signal we observe, $\hat{x}$ is the estimate of the target $x$, $\{f_j\}$ are $J$ first-order derivative filters and $\beta$ is a weight parameter. Setting $\gamma$ to one, with the corresponding filters, gives the special case of total-variation methods~\cite{rudin1992nonlinear}. Generally these approaches are an optimization of the form
\begin{equation}
	\hat{x} = \argmin_x \psi_d(x,y) + \beta \cdot \psi_p(\mathcal{T}(x)),
\end{equation}
where $\psi_d$ is the data-fidelity loss term, and $\psi_p$ incorporates the prior information on a transformation $\mathcal{T}$ of $x$ that could be the identity. $\mathcal{T}$ can also be based on derivatives~\cite{krishnan2009fast}, or wavelet~\cite{sardy2001robust} and other sparsifying transformations. For instance, \medits{Weighted Nuclear Norm Minimization (WNNM)}~\cite{WNNM} assumes that subsets of similar image patches are low-rank and uses a weighted nuclear norm for the low-rank minimization problem on similar patch groups. As with many classic image denoising methods, WNNM adapts $\beta$ based on the noise level and controls the data fidelity weight as such. However, as we noted in our introduction, these methods face two shortcomings. First, $\beta$ is not adapted based on the confidence in the prior given the degraded observation, but only on the quality of the latter. Hence, it is adapted based on the signal quality, such as the noise level, but also only following certain heuristics. Second, the prior itself is fixed based on hand-designed heuristics. We preserve an interpretable control over the contribution of the prior and decouple it from data fidelity, and we exploit learned network priors and increasing the flexibility in the fusion weight. Therefore, this weight is \textit{learned} in our framework and can adapt both to the quality of the observed data, as well as the fitness of the prior, given the test observation.

\subsection{Deep Neural Networks}
\medits{Convolutional neural networks are able to learn rich image priors. These learned priors have improved image restoration results for various applications~\cite{jain2009natural,dncnn,zhang2016colorful,yu2019free}}. These methods use sample-based learning and can extract prior information from large image datasets. \medits{Further improvements are achieved with discriminative learning, notably discriminators that take into account the image degradation model}~\cite{pan2020physics}.
This has enabled these deep learning methods to improve the state of the art on many restoration tasks~\cite{zhang2019residual}. However, the learned priors are implicit, meaning neither the prior nor its contribution can be disentangled from the data fidelity component in the final restored output. \meditstwo{In a concurrent paper, Dong \textit{et al.} propose to learn the data fidelity term and the prior term with separate neural network components~\cite{dong2021learning}. The learning, and the final restoration output, are however carried out within a unified optimization. This makes the disentanglement of the two components and their relative contribution weights not possible downstream in the spatial domain.} As recently shown for super-resolution and denoising tasks~\cite{elhelou2020stochastic}, neural networks can learn a frequency-conditional hallucination that is prone to overfitting to the training degradation models. Another recent example is in 3D reconstruction~\cite{tatarchenko2019single}, where networks learn to recognize observations and to use memorized data samples, rather than to perform the reconstruction. In other words, the prior contribution can dominate over the data fidelity. \meditstwo{Despite its importance, controlling this trade-off is, however, not attainable within the neural-network-based solutions.} Our proposed framework enables us to exploit the strength of learned network priors and keep both control and insight over the data fidelity and prior trade-offs.

Extracting prior information from neural networks is possible through an inversion process~\cite{donahue2017adversarial,albright2019source}. By searching the network-learned space of image distributions, it is possible to project on it in a fashion similar to dictionary-based methods. Generative networks are sufficiently powerful to be trained to learn different distributions, such as image or noise distributions~\cite{chen2018image}. A network inversion is carried out in~\cite{gu2020image}, where it is used for different image processing tasks. We discuss this in more detail in the following section. A generative network inversion is also conducted in~\cite{pan2020dgp}. However, the method performs a fine-tuning of the pre-trained generator that goes against our objective to project on a fixed learned space. We also emphasize that our goal is not to improve such priors, rather to use them in our framework as image-projection spaces.

\subsection{Signal Adaptation of Priors}
As described in our discussion on classic methods, some of them~\cite{WNNM} adapt the weight assigned to their prior term according to the quality of the observed signal. However, the fitness of the chosen prior can itself be image dependent. In other words, the prior can be accurate on certain images, but not as fit to be applied to others. Yet this is rarely accounted for in the literature. In the content-aware image prior presented in~\cite{cho2010content}, although the weight of the prior itself is not adaptive, the hyper-Laplacian prior used is tweaked to adjust to the texture in the observed signal. Similarly, the method in~\cite{choi2018fast} carefully selects its filters upon processing of the observed signal, hence altering its implicit priors. Also in the same spirit, some recent deep learning methods have tried to adapt to the observed inputs, through self-supervised weight modification~\cite{lee2020self}, or novel learning~\cite{bau2020semantic}. This approach has even appeared in recent classification work to adjust to distribution shifts~\cite{sun2020test}. Such methods address the issue of the fitness of the prior to the given input by modifying the former on the fly. However, once a prior is selected, its fitness relative to the observed signal's quality is dismissed. The weight of the prior term is therefore not adaptive, and the prior's contribution cannot be decoupled from data fidelity.

\section{Method} \label{sec:bigp_method}
In designing our method, we address the shortcomings discussed in the introduction. We present a framework where the prior and the data-fidelity terms are explicit. This enables us to exploit the modeling strength of deep neural networks for the prior and enables us to learn a weight between the prior and the data fidelity that is doubly adaptive to the quality of the observation and to the fitness of the prior to the input's distribution. Rather than combining the contribution of the prior and the data-fidelity terms through an optimization, we explicitly enforce their fusion in the final output. This explicit decoupling of the two terms enables us, as well as downstream applications, to gain in restoration interpretability. In this section, we present the mathematical details of our proposed method and its relation to classic families of restoration algorithms. We also present a network-based prior that relies on generative-network projection and introduce our approach for learning the adaptive weight without guided supervision.

\subsection{Mathematical Formulation} \label{sec:bigp_method_math}
Given an observed signal $y$ that is a degraded version of the image $x$, our restoration estimate $\hat{x}$ is formulated as
\begin{equation} \label{eq:bigp_formulation}
	\hat{x} = \underbrace{\left(1-\phi(y;\theta_1)\right) \odot g^{-1}(y)}_{data~fidelity} + \underbrace{\phi(y;\theta_1) \odot G(z^*;\theta_2)}_{prior},
\end{equation}
where $g^{-1}(\cdot)$ is a bijective function that we discuss in what follows, $\phi(\cdot;\theta_1)$ is an estimator for the fusion factor, parameterized by $\theta_1$, and that assigns adaptive weights to the prior-based hallucination and the data fidelity. It is a generalization of $\beta$ that we learn internally from sample-based training. $G(z^*;\theta_2)$ is the prior-based hallucination, parameterized by $\theta_2$, described in detail in the following, and $\odot$ is the pixel-wise multiplication operator.
To ensure a very strict lossless data-fidelity term, we restrict $g^{-1}(\cdot)$ to the set of bijective functions. We can choose it such that $g(\cdot)$ is close to the degradation model $f(\cdot)$ of the restoration task, as described in our experimental setup. We note that this formulation is closely related to the classic restoration methods based on explicit prior optimizations discussed in our related work. The difference is that our prior is based on a trainable neural network $G$, and that our fusion factor is also learned to be adaptive, per sample, both to the quality of the observed data and to the fitness of the prior.

We present the \textbf{relation to MAP estimation} in connection with the work in~\cite{el2020blind}. The authors derive a MAP estimate for \medits{Additive White Gaussian Noise (AWGN)} removal. \medits{This MAP estimate is the mode of the posterior probability distribution. The solution is derived} where the additive noise ($y_i=x_i+n_i$) follows the normal distribution $\mathcal{N}(0,\sigma_n)$, and an explicit image prior is enforced. More precisely, the solution is derived with the assumption of a Gaussian prior~\cite{romdhani2005estimating} \textit{on the pixel distribution}. With this model, the prior distribution for a pixel value $x_i$ follows $\mathcal{N}(\Bar{x}_i,\sigma_{x_i})$ \medits{with mean $\Bar{x}_i$ and standard deviation $\sigma_{x_i}$}, and this yields a MAP estimate
% \begin{equation}
%     P_{X_i}(x_i) = \frac{1}{\sqrt{2\pi\sigma_x^2}}e^{-\frac{(x_i-\Bar{x_i})^2}{2\sigma_x^2}},
% \end{equation} 
\begin{equation}
	\hat{x}_i = \argmax_{x_i} P_{X_i|Y_i}(x_i|y_i) = \frac{y_i}{1+1/S_i} + \frac{\Bar{x}_i}{1+S_i},
\end{equation}
with $S_i$ being the signal-to-noise ratio defined as
\begin{equation}
	S_i\triangleq \frac{\sigma_{x_i}^2}{\sigma_n^2}.    
\end{equation}
Note how $S_i$ is, in fact, dependent on signal quality (\medits{through the noise standard deviation} $\sigma_n$), as well as the confidence in the prior (\medits{through the pixel standard deviation} $\sigma_{x_i}$). Indeed, intuitively the larger $\sigma_{x_i}$ is, the less reliable the prior term $\Bar{x}_i$ is; and the smaller it is, the more reliable the prior term is. In this special case of our general formulation,
\begin{equation}
	\phi(y_i)=\frac{1}{{1+S_i}},    
\end{equation}
$g(\cdot)$ is the identity mapping, and the prior is the expected value over the distribution of the input $\mathbb{E}_{X_i}[x_i]$. Our formulation in Equation~\eqref{eq:bigp_formulation} generalizes this solution to non-Gaussian, as well as image-wise prior distributions, while taking into account signal quality and prior confidence. 

We also describe the \textbf{relation to dictionary-based methods}. Dictionary-based methods~\cite{giryes2014sparsity,rubinstein2012analysis} generally follow the formulation
\begin{equation}
	\hat{x} = \argmin_{x, d(x, Dv)<\epsilon} \psi_d(x,y) + \beta \cdot \psi_p(v),
\end{equation}
where $D$ is the dictionary, specifically, a vector set that spans the dictionary space, $v$ holds the coordinates of a point in that space, $d(\cdot,\cdot)$ is a distance function, and $\epsilon$ is a small value in $\mathbb{R}_+$. It is typical to use a $\psi_p$ that encourages sparsity, thus to assume that the dictionary captures the main directions of variation in an image. This sparsity of $v$ parallels restrictions on the generative latent space.
Effectively, enforcing
\begin{equation}
	d(x, Dv)<\epsilon    
\end{equation}
is a subtle relaxation of the constraint $x \in \operatorname{span}(D)$, which enforces the prior assumption that the image must belong to the dictionary space. This would correspond in our formulation of Equation~\eqref{eq:bigp_formulation} to $x \in \operatorname{span}(G)$, where in our case the dictionary space is instead the learned space of a generative network. In our formulation, the restriction is enforced only on our decoupled prior element, rather than having to enforce it on $x$ itself and then relaxing it through a tweaking of $\epsilon$. 

In summary, our formulation can be viewed as a general framework of MAP estimation and as a generalization of dictionary-based methods. We choose a strict data-fidelity term that preserves a bijective relation to the observed signal, and a fusion factor that takes into account both signal quality and prior confidence. The following two sections discuss in more detail the prior term and the $\phi$ fusion factor learning.

\begin{table}[]
	\centering
	\begin{tabular}{lcc}
		\toprule
		& $\phi$ explicitly known & $\phi$ relation to \textit{data fidelity} \\ \cline{1-3}
		Colorization & \ding{55} & Luminance and edge related \\
		Inpainting   & \ding{51} & Binary mask based \\
		Denoising    & \ding{55} & Noise-level adaptive \\\bottomrule \\
	\end{tabular}
	\caption[Overview of $\phi$ properties across restoration tasks]{The $\phi$ map values are only explicitly known for inpainting, but are always related to the data-fidelity and prior-confidence terms discussed in our mathematical formulation. Indeed, in colorization there exist strong relations between luminance and the fidelity of the observed data, in inpainting this directly matches the applied mask, and in denoising the noise level determines the fidelity of the observation. The $\phi$ map also, across all tasks, depends on the confidence in the prior.}
	\label{table:tasks_phi_relation}
\end{table}

\begin{table}[t!]
	\centering
	\begin{tabular}{lcc}
		\toprule
		& Bedroom set & Church set \\
		Method & AuC~\cite{zhang2016colorful} $\uparrow$ & AuC~\cite{zhang2016colorful} $\uparrow$ \\ \cline{1-3}
		\rowcolor{Gray}
		Colorful colorization~\cite{zhang2016colorful} & \underline{88.55} & 89.13 \\
		Deep image prior~\cite{ulyanov2018deep} & 84.33 & 83.31 \\
		Feature map opt.~\cite{bau2020semantic} & 85.41 & 86.10 \\
		mGAN prior~\cite{gu2020image} & 88.52 & \underline{89.69} \\
		Ours & \textbf{89.27} & \textbf{90.64} \\\bottomrule \\
	\end{tabular}
	\caption[Quantitative colorization results]{Quantitative AuC (\%) results for image colorization on the Bedroom and Church test sets. The higher the value is, the lower the cumulative colorization error curve is. We highlight, with background shaded in gray, the widely used task-specific supervised method. The best results are shown in bold, and the second best are underlined.}
	\label{table:colorization_results}
\end{table}

% Bedroom (263) Church (260)
\begin{figure*}[t!]
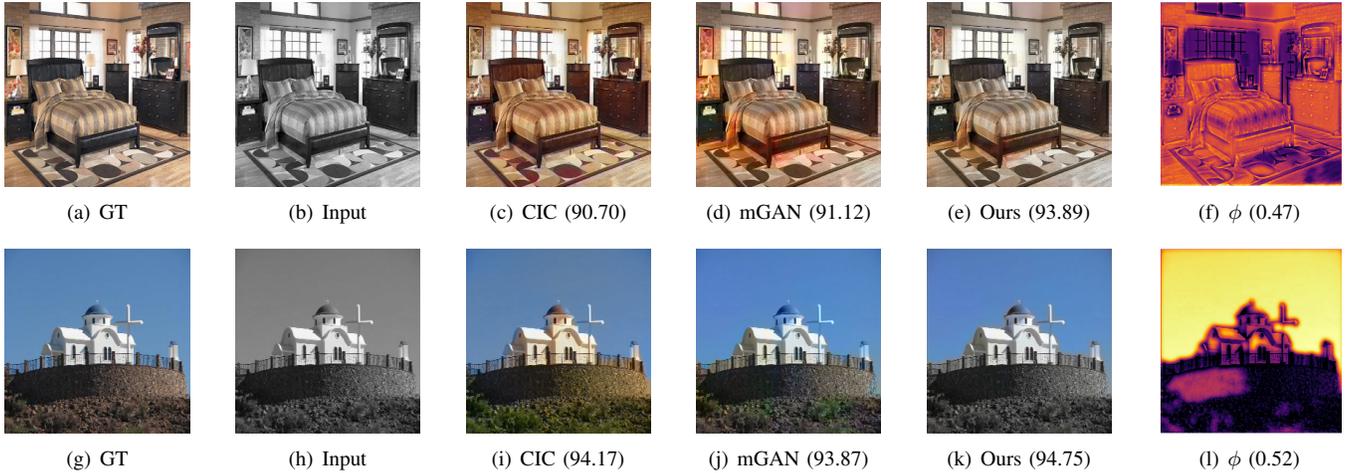

	\centering
	\subfigBIGP{GT}{Colorization/263_gt.png}
	\subfigBIGP{Input}{Colorization/263_inp.png}
	\subfigBIGP{CIC (90.70)}{Colorization/263_cic.png}
	\subfigBIGP{mGAN (91.12)}{Colorization/263_mgan.png}
	\subfigBIGP{Ours (93.89)}{Colorization/263.png}
	\subfigBIGPphi{$\phi$ (0.47)}{Colorization/263_phi.png}
	
	\subfigBIGP{GT}{Colorization/260_gt.png}
	\subfigBIGP{Input}{Colorization/260_inp.png}
	\subfigBIGP{CIC (94.17)}{Colorization/260_cic.png}
	\subfigBIGP{mGAN (93.87)}{Colorization/260_mgan.png}
	\subfigBIGP{Ours (94.75)}{Colorization/260.png}
	\subfigBIGPphi{$\phi$ (0.52)}{Colorization/260_phi.png}

	\caption[Visual colorization results, with our $\phi$ map]{From left to right are the ground-truth image (GT), the grayscale input, the results of colorful image colorization (CIC)~\cite{zhang2016colorful}, mGAN~\cite{gu2020image}, and ours, with the AuC (\%), and our channel-averaged $\phi$ map (with global average between parentheses). The darker colors indicate values of $\phi$ closer to 0, whereas bright yellow indicates those closer to 1.} \label{fig:colorization}
\end{figure*}

\subsection{Generative-Space Projection Prior}
Theoretically, an inference method can be used to replace the prior term. For instance, a feed-forward network's output can replace $G(z^*)$ in Equation~\eqref{eq:bigp_formulation}. However, such a network trained with supervision takes into account both the data-fidelity and prior terms, albeit without any insight as to how much prior-based hallucination occurs or any control over the different contributions. Therefore, in order to best decouple data fidelity from prior hallucination, we opt for a pre-trained generative network inversion to act as the learned prior. Effectively, this is a better strategy for decreasing the upper bound on a worst-case hallucination contribution. The inversion produces a sampling from the generative space, or a projection on that space as in dictionary-based projections discussed earlier. The latent code $z^*$ for the generative-space projection is obtained as
\begin{equation} \label{eq:general_inversion}
	z^* = \argmin_z \mathcal{L}_G(f(G(z)),y),
\end{equation}
where $\mathcal{L}_G$ can be a weighted average of $\ell_1$, $\ell_2$, and perceptual losses, and $f(\cdot)$ is the degradation model of a restoration task. When using a single latent code, very limited information can be encoded, which yields a coarse prior, notably for high-resolution images. To avoid this loss of expressiveness, we use the recent multi-code \medits{Generative Adversarial Network (GAN)} inversion method that splits the generative network $G$ into two stages, at layer $l$~\cite{gu2020image}. The first stage $G_1^{(l)}$ generates multiple feature space representations, each corresponding to one of $N$ latent codes $\{z_n^*\}_{n=1}^N$, where every $\alpha$ is a vector of length equal to the number of feature-space channels. The second stage $G_2^{(l)}$ generates the output image based on a fused feature representation by using adaptive channel weights $\{\alpha_n^*\}_{n=1}^N$. The latent codes and adaptive weights are obtained, as in Equation~\eqref{eq:general_inversion}, by an inversion optimization
\begin{equation}
	\{z_n^*\}_{n=1}^N, \{\alpha_n^*\}_{n=1}^N = \argmin_{\{z_n\}_{n=1}^N, \{\alpha_n\}_{n=1}^N} 
	\mathcal{L}(f(x^{inv}),x),
\end{equation}
where the inverted image $x^{inv}$ is given by
\begin{equation}
	G(z;\alpha,\theta_2) \triangleq x^{inv} = G_2^{(l)}\left( \sum_{n=1}^{N} G_1^{(l)}(z_n) \cdot \alpha_n \right).
\end{equation}
Our image prior term in Equation~\eqref{eq:general_inversion} is then given by $G(z^*;\alpha^*,\theta_2)$, where $\theta_2$ are the frozen weights of the generative sub-networks $G_1$ and $G_2$. We also note that randomly traversing the latent space of a generative network can potentially produce hallucinated images that lie outside the natural image manifold~\cite{menon2020pulse}. This is averted by the guided inversion loss that maps the generative output, through the degradation model, to the observed image. The case of the generative projection being outside the natural-image manifold, which can occur when the degradation is extreme, still does not pose an issue in our framework. Indeed, this projection is already treated in our approach as a prior hallucination that might not be faithful to the original image.

\subsection{Guide-Free $\phi$ Learning}
A guided learning of the parameters $\theta_1$ to predict $\phi$ is possible for a task such as inpainting but impossible for other tasks. This is simply because inpainting is the extreme case where signal quality is binary, specifically zero at the masked areas. For other tasks, a target $\phi$ cannot be readily obtained. We thus train a network with weights $\theta_1$ to predict $\phi$ in an end-to-end manner, with $\phi$ effectively being an intermediate feature space having no explicit learning loss. Our mini-batch training loss $\mathcal{L}(x,y;\theta_1)$ for learning $\theta_1$ is given by (we use $\phi$ to also denote the network outputting it, for better readability)
% \begin{equation} \label{eq:bigp_loss}
%     || \left(1-\phi(y)\right) \odot g^{-1}(y) + \phi(y) \odot G(z^*) - x||_2^2 + \rho \cdot ||\phi(y)||_1,
% \end{equation}
\begin{equation} \label{eq:bigp_loss}
	\begin{multlined}
		\mathcal{L}(x,y;\theta_1,\theta_2) = || \left(1-\phi(y,\theta_1)\right) \odot g^{-1}(y) \\
		+ \phi(y,\theta_1) \odot G(z^*;\alpha^*,\theta_2) - x||_2^2 + \rho \cdot ||\phi(y,\theta_1)||_1,
	\end{multlined}
\end{equation}

where $\rho$ is a scalar weight that we discuss next, and the parameters $\theta_2$ of the generative network are the frozen weights of a \textit{pre-trained} generative network. This end-to-end training enables the network predicting $\phi$ to learn to assess, based on the observation $y$, the quality of that observed image, as well as the fitness of the prior to this observation.

\begin{table}[t]
	\centering
	\begin{tabular}{lccc}
		\toprule
		Method & PSNR $\uparrow$ & SSIM $\uparrow$ & LPIPS $\downarrow$ \\ \cline{1-4}
		\rowcolor{Gray}
		DeepFill v2~\cite{yu2018generative,yu2019free} & \textbf{26.56} & \textbf{0.9555} & \textbf{0.0191} \\
		Feature map opt.~\cite{bau2020semantic} & 14.75 & 0.4563 & - \\
		Deep image prior~\cite{ulyanov2018deep} & 17.92 & 0.4327 & -\\
		mGAN prior~\cite{gu2020image} & 20.55 & 0.5823 & 0.2070 \\
		Ours & \underline{25.32} & \underline{0.9240} & \underline{0.0376}\\
		\bottomrule \\
	\end{tabular}
	\caption[Quantitative central inpainting results]{Quantitative PSNR ($dB$), SSIM, and LPIPS results for central image inpainting. We mask out a $64\times 64$ patch from the center of each input image. The task-specific state-of-the-art method is highlighted with background shaded in gray. The best results are shown in bold, and the second best are underlined.}
	\label{table:inpcrop_results}
\end{table}

\begin{table}[t]
	\centering
	\begin{tabular}{clccc}
		\toprule
		Test & Method & PSNR $\uparrow$ & SSIM $\uparrow$ & LPIPS $\downarrow$ \\ \cline{1-5}
		\parbox[t]{2mm}{\multirow{2}{*}{\rotatebox[origin=c]{90}{Bed.}}} & mGAN prior~\cite{gu2020image} & 20.34 & 0.5902 & 0.2134 \\
		& Ours & \textbf{23.22} & \textbf{0.8598} & \textbf{0.0775}\\ \hdashline
		\parbox[t]{2mm}{\multirow{2}{*}{\rotatebox[origin=c]{90}{Chu.}}} & mGAN prior~\cite{gu2020image} & 19.33 & 0.5359 & 0.2235 \\
		& Ours & \textbf{21.94} & \textbf{0.8509} & \textbf{0.0855}\\ \hdashline
		\parbox[t]{2mm}{\multirow{2}{*}{\rotatebox[origin=c]{90}{Conf.}}} & mGAN prior~\cite{gu2020image} & 19.38 & 0.5641 & 0.2062 \\
		& Ours & \textbf{22.20} & \textbf{0.8318} & \textbf{0.0785}\\
		\bottomrule \\
	\end{tabular}
	\caption[Quantitative randomized-inpainting results]{Quantitative PSNR ($dB$), SSIM, and LPIPS results for randomized-masking inpainting on the Bedroom, Church (Outdoor), and Conference test sets. The randomized masking increases the difficulty of predicting our $\phi$ maps. To analyze the effect of mask randomization on the performance of our $\phi$ prediction compared to the central inpainting task, we compare the prior-based results to ours.}
	\label{table:inpvar_results}
\end{table}

%INPCROP: bedroom (250 -254- Fail: 289) conference (-254- 296)
%INPVAR: bedroom(263,293F, )
\begin{figure*}[t]
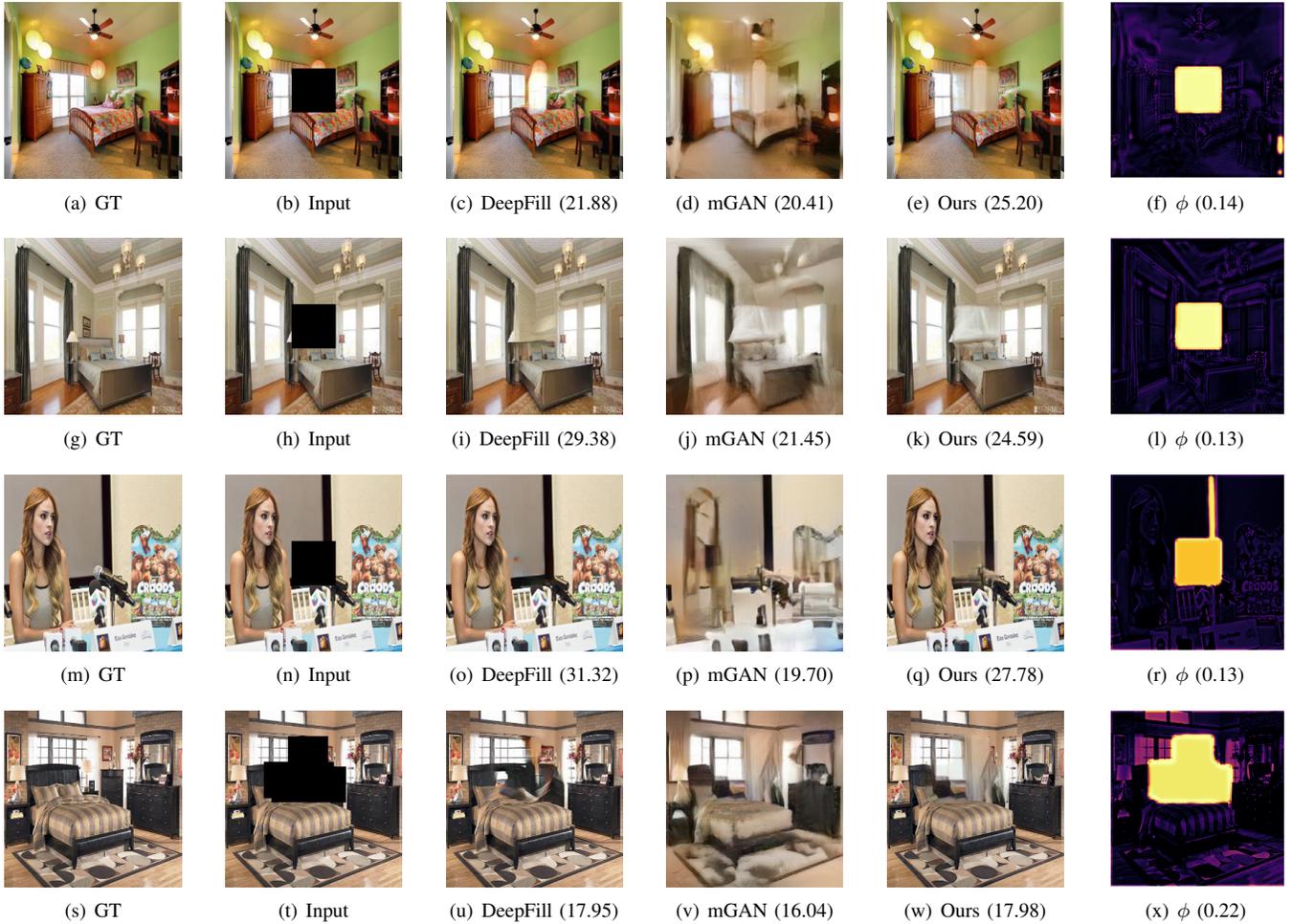

	\centering
	\subfigBIGP{GT}{Inpainting/250_gt.png}
	\subfigBIGP{Input}{Inpainting/250_inp.png}
	\subfigBIGP{DeepFill (21.88)}{Inpainting/250_dfill.png}
	\subfigBIGP{mGAN (20.41)}{Inpainting/250_mgan.png}
	\subfigBIGP{Ours (25.20)}{Inpainting/250.png}
	\subfigBIGPphi{$\phi$ (0.14)}{Inpainting/250_phi.png}
	
	\subfigBIGP{GT}{Inpainting/289_gt.png}
	\subfigBIGP{Input}{Inpainting/289_inp.png}
	\subfigBIGP{DeepFill (29.38)}{Inpainting/289_dfill.png}
	\subfigBIGP{mGAN (21.45)}{Inpainting/289_mgan.png}
	\subfigBIGP{Ours (24.59)}{Inpainting/289.png}
	\subfigBIGPphi{$\phi$ (0.13)}{Inpainting/289_phi.png}
	
	 	\subfigBIGP{GT}{Inpainting/296_gt.png}
	 	\subfigBIGP{Input}{Inpainting/296_inp.png}
	 	\subfigBIGP{DeepFill (31.32)}{Inpainting/296_dfill.png}
	 	\subfigBIGP{mGAN (19.70)}{Inpainting/296_mgan.png}
	 	\subfigBIGP{Ours (27.78)}{Inpainting/296.png}
	 	\subfigBIGPphi{$\phi$ (0.13)}{Inpainting/296_phi.png}
	
	\subfigBIGP{GT}{VarInp/263_gt.png}
	\subfigBIGP{Input}{VarInp/263_inp.png}
	\subfigBIGP{DeepFill (17.95)}{VarInp/263_dfill.png}
	\subfigBIGP{mGAN (16.04)}{VarInp/263_mgan.png}
	\subfigBIGP{Ours (17.98)}{VarInp/263.png}
	\subfigBIGPphi{$\phi$ (0.22)}{VarInp/263_phi.png}
	
	\caption[Visual central and randomized inpainting results, with our $\phi$ map]{From left to right are the ground-truth image (GT), the masked input, the results of DeepFill v2~\cite{yu2018generative,yu2019free}, mGAN~\cite{gu2020image}, and ours, with the PSNR in $dB$, and our channel-averaged $\phi$ map (with global average between parentheses). The first three rows show example images from the standard central-inpainting benchmark, and the last row is an example from our randomized-inpainting experiment.} \label{fig:inp}
\end{figure*}

\medits{\textbf{Fidelity-Bias Balance.}} For certain image test cases, the quality of the data-fidelity term can be very similar to that of the learned prior, at least over some subsets of pixels. This would induce no change in the loss term for varying values of our fusion factor $\phi$, as all would result in similar final outputs. However, for these cases, it is not necessary to \textit{hallucinate} information as the data fidelity is also just as accurate. \medits{By hallucinated information, we mean that which is not in direct relation with the observed data, but rather comes from previously learned priors.} Therefore, we address these edge cases by adding an auxiliary loss on the $\ell_1$ norm of $\phi$ in Equation~\eqref{eq:bigp_loss}, which can additionally regularize the feature learning process~\cite{el2020al2}. This term enforces that the training favors smaller values of $\phi$ such that the overall contribution of the data fidelity term is maximized when this is not detrimental to the quality of the final output. This fidelity-bias term is weighted by the scalar $\rho$ in Equation~\eqref{eq:bigp_loss}.

% Bedroom (282, 283, 293) Conference (268), Failure case conference (251, 269)
\begin{figure*}[t!]
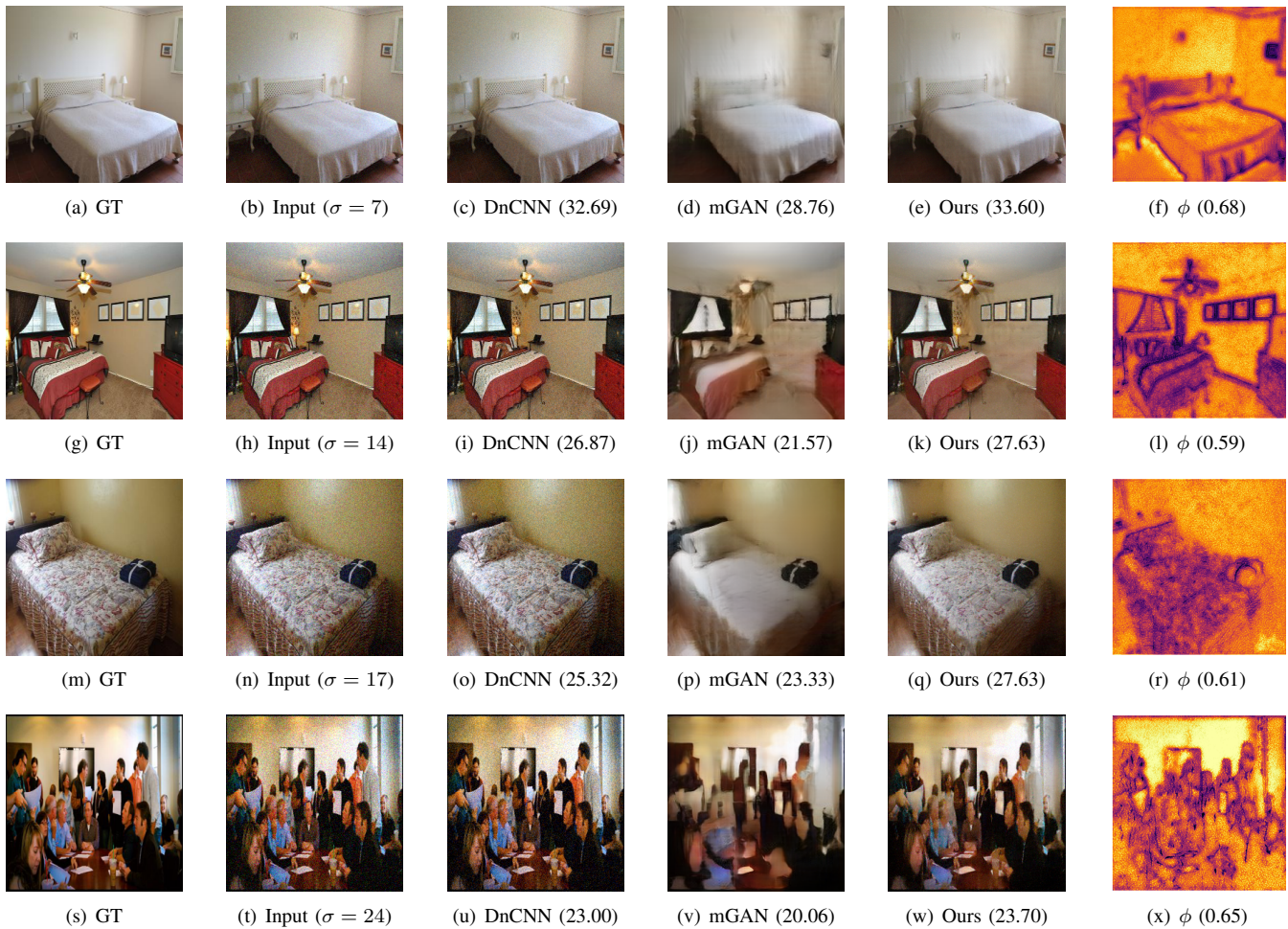

	\centering
	\subfigBIGP{GT}{AWGN/282_gt.png}
	\subfigBIGP{Input ($\sigma=7$)}{AWGN/282_inp.png}
	\subfigBIGP{DnCNN (32.69)}{AWGN/282_dncnn.png}
	\subfigBIGP{mGAN (28.76)}{AWGN/282_mgan.png}
	\subfigBIGP{Ours (33.60)}{AWGN/282.png}
	\subfigBIGPphi{$\phi$ (0.68)}{AWGN/282_phi.png}
	
	\subfigBIGP{GT}{AWGN/283_gt.png}
	\subfigBIGP{Input ($\sigma=14$)}{AWGN/283_inp.png}
	\subfigBIGP{DnCNN (26.87)}{AWGN/283_dncnn.png}
	\subfigBIGP{mGAN (21.57)}{AWGN/283_mgan.png}
	\subfigBIGP{Ours (27.63)}{AWGN/283.png}
	\subfigBIGPphi{$\phi$ (0.59)}{AWGN/283_phi.png}
	
	\subfigBIGP{GT}{AWGN/293_gt.png}
	\subfigBIGP{Input ($\sigma=17$)}{AWGN/293_inp.png}
	\subfigBIGP{DnCNN (25.32)}{AWGN/293_dncnn.png}
	\subfigBIGP{mGAN (23.33)}{AWGN/293_mgan.png}
	\subfigBIGP{Ours (27.63)}{AWGN/293.png}
	\subfigBIGPphi{$\phi$ (0.61)}{AWGN/293_phi.png}
	
	\subfigBIGP{GT}{AWGN/268_gt.png}
	\subfigBIGP{Input ($\sigma=24$)}{AWGN/268_inp.png}
	\subfigBIGP{DnCNN (23.00)}{AWGN/268_dncnn.png}
	\subfigBIGP{mGAN (20.06)}{AWGN/268_mgan.png}
	\subfigBIGP{Ours (23.70)}{AWGN/268.png}
	\subfigBIGPphi{$\phi$ (0.65)}{AWGN/268_phi.png}
	
	\caption[Visual AWGN denoising results, with our $\phi$ map]{From left to right are the ground-truth image (GT), the noisy input with the AWGN standard deviation, the results of DnCNN~\cite{dncnn}, mGAN~\cite{gu2020image}, and ours, with the PSNR in $dB$, and our channel-averaged $\phi$ map (with global average between parentheses).} \label{fig:denblind}
\end{figure*}

\section{Experiments}
We conduct experiments on image colorization, inpainting, and blind AWGN removal. \medits{When comparing to other methods in our experiments, we use the setups recommended by the authors in their original papers and public codes.} For Colorization does not induce an explicit solution for $\phi$, aside for certain exceptions that we discuss in the next section, such as edges and extreme luminance areas. Inpainting induces an explicitly known solution for $\phi$. Whereas, AWGN does not have an explicit solution for $\phi$, as the image prior is not explicitly formulated. However, the AWGN experimental setup enables us to analyze the guide-free learning of $\phi$, which would intuitively fluctuate mainly with the noise level (direct relation), but also marginally with the uncertainty in the prior (opposite relation), as described in our discussion section. This is summarized in Table~\ref{table:tasks_phi_relation} and analyzed in the following sections.

\subsection{Experimental Setup} \label{sec:sec:bigp_expsetup}
As described in Section~\ref{sec:bigp_method}, we use the multi-code GAN inversion approach\footnote{mGAN~\cite{gu2020image}: \url{https://github.com/genforce/mganprior}} for our generative-space projection prior. The pre-trained GAN models, which correspond to each dataset used, are all different versions of the \medits{Progressive Growing of GANs (PGGAN)}~\cite{karras2018progressive} network\footnote{PGGAN~\cite{karras2018progressive}: \url{https://github.com/tkarras/progressive_growing_of_gans}}. They are pre-trained on the Bedroom, Church (Outdoor), and Conference room datasets taken from the LSUN database~\cite{yu2015lsun}. The details for each experiment follow the settings given by the authors of~\cite{gu2020image} and are given in the following sections. We note that any generative network, such as DCGAN~\cite{radford2015unsupervised}, LR-GAN~\cite{yang2017lr}, CVAE-GAN~\cite{bao2017cvae}, StyleGAN~\cite{karras2019style}, StyleGAN2~\cite{karras2020analyzing}, or even any future method allowing projections or sampling from a learned image distribution, can be used for the projection prior of our method. To enable direct comparisons with with mGAN~\cite{gu2020image}, we use the PGGAN in our experiments. We use AuC~\cite{deshpande2015learning,zhang2016colorful}, PSNR, SSIM~\cite{wang2004image}, and the perceptual metric LPIPS~\cite{zhang2018unreasonable} in our quantitative evaluations.

For our fusion factor learning, we train the same backbone network with the same settings for all of our experiments. The architecture is inspired by~\cite{dncnn} and is a residual learning made up of a sequence of convolutional, batch normalization, and ReLU blocks. We omit further architecture details that can be found in our code. We use a batch size of 8 \medits{except for the real denoising experiment where the batch size is 4}, a starting learning rate of 0.01, and a fidelity-bias balancing weight $\rho=1e-5$. \medits{A very small balancing weight is sufficient ($1e-5$), because our objective is to favor fidelity over hallucination only when they are deemed to have the same accuracy.} We train for 25 epochs with random shuffling and update the learning rate following a cosine annealing with warm restarts scheduler~\cite{loshchilov2017sgdr}. The restart period is adaptive to the batch size such that it is always 4 epochs. We also note for reproducibility that training with images that are normalized to $[0,1]$ and then zero-centered is empirically observed to improve the final results. The same normalization is then performed before inference and inverted once the output is obtained. We train our model with the loss of Equation~\eqref{eq:bigp_loss} on a subset of the LSUN validation set that corresponds to each of the large training sets used for pre-training the PGGAN models, and we test on the remaining subset. %We present in our Supplementary Material the same experiments repeated on train-test subsets taken from the training set of the pre-trained PGGAN, which gives an advantage to the mGAN inversion~\cite{gu2020image} as it is tested on images seen by its pre-trained GAN.

\begin{table}[t]
	\centering
	\begin{tabular}{clccc}
		\toprule
		Test & Method & PSNR $\uparrow$ & SSIM $\uparrow$ & LPIPS $\downarrow$ \\ \cline{1-5}
		\parbox[t]{2mm}{\multirow{3}{*}{\rotatebox[origin=c]{90}{Bed.}}} & \cellcolor{Gray}DnCNN$^\dagger$~\cite{dncnn} & \cellcolor{Gray}24.96 & \cellcolor{Gray}0.5804 & \cellcolor{Gray}0.1859 \\
		& mGAN prior~\cite{gu2020image} & 22.72 & 0.6257 & 0.1978 \\
		& Ours & \textbf{26.80} & \textbf{0.7279} & \textbf{0.0998}\\ \hdashline
		\parbox[t]{2mm}{\multirow{3}{*}{\rotatebox[origin=c]{90}{Church}}} & \cellcolor{Gray}DnCNN$^\dagger$~\cite{dncnn} & \cellcolor{Gray}22.40 & \cellcolor{Gray}0.5166 & \cellcolor{Gray}0.2046 \\
		& mGAN prior~\cite{gu2020image} & 21.12 & 0.5643 & 0.2065 \\
		& Ours & \textbf{23.38} & \textbf{0.5959} & \textbf{0.1435}\\ \hdashline
		\parbox[t]{2mm}{\multirow{3}{*}{\rotatebox[origin=c]{90}{Conf.}}} & \cellcolor{Gray}DnCNN$^\dagger$~\cite{dncnn} & \cellcolor{Gray}22.81 & \cellcolor{Gray}0.5310 & \cellcolor{Gray}0.2167 \\
		& mGAN prior~\cite{gu2020image} & 21.49 & 0.5962 & 0.1968 \\
		& Ours & \textbf{24.70} & \textbf{0.6578} & \textbf{0.1192}\\
		\bottomrule \\
	\end{tabular}
	\caption[Quantitative blind denoising results]{PSNR ($dB$), SSIM, and LPIPS results for AWGN removal on the Bedroom, Church, and Conference sets. The noise follows a Gaussian distribution with standard deviation sampled uniformly at random from [5,50] per image. $^\dagger$We retrain and test DnCNN with the same data and setup as ours.}
	\label{table:denblind_results}
\end{table}

\subsection{Colorization}
For the colorization of a grayscale input image, unlike inpainting for example, it is much less predictable what an ideal $\phi$ map would be. We conduct colorization experiments, where the grayscale input is the luminance channel, and we evaluate the error on the $ab$ color space. The AuC metric~\cite{deshpande2015learning,zhang2016colorful} computes the area under the cumulative percentage $\ell_2$ error distribution curve in the $ab$ space. The percentage is that of pixels lying within an error threshold that is swept over $[0,150]$ in steps of one. For generative network inversion, we use the sixth layer of the PGGAN for the feature composition, with 20 latent codes, and $\ell_2$ and VGG-16 perceptual loss~\cite{simonyan2015very}, optimized with gradient descent for 1500 iterations, following~\cite{gu2020image}. Our $g^{-1}(y)$ function duplicates the grayscale channel over each of the color channels. The remaining details follow the experimental setup of Section~\ref{sec:sec:bigp_expsetup}.

We present our quantitative image colorization results in Table~\ref{table:colorization_results}, along with those of the deep image prior~\cite{ulyanov2018deep}, the feature map optimization~\cite{bau2020semantic}, the colorful image colorization~\cite{zhang2016colorful}, which is a feed-forward method supervised specifically for colorization, and the mGAN prior~\cite{gu2020image}. We note the considerable improvement of our method, despite the restriction of enforcing a strict data fidelity. \medits{Relative to the mGAN results, we improve by +0.75AuC on the Bedroom set reaching 89.27AuC, and by +0.95AuC on the Church set reaching 90.64AuC. These results even exceed the performance of the task-specific colorful image colorization method~\cite{zhang2016colorful} on the two test sets, by +0.72AuC and +1.51AuC, respectively.}

Visual results are shown in Figure~\ref{fig:colorization} for the different image colorization methods. We can observe that $\phi$ is lower on image edges, which indeed generally constitute information that is not lost by the grayscale degradation. We observe as well that $\phi$ tends to be low when the luminance is around extreme values, as in such cases the grayscale images are faithful to the original color images. In both of these cases, it is the confidence in the data fidelity that is adapted to the observation. We also note, for instance in the sample of the second row in Figure~\ref{fig:colorization}, that $\phi$ can be very insightful. It indicates that the color of the sky was heavily hallucinated, whereas the bottom half and the church dome use almost no prior hallucination. This is advantageous for downstream tasks as the dome was, in fact, incorrectly hallucinated by the generative-network projection prior. This is similar for the grass \medits{part of that same image in the second row. The generative prior incorrectly adds a green color for the grass as can often be expected. This hallucination is however deemed incorrect by our $\phi$ map, i.e., the added fake color is not correct. Therefore, our final output relies more heavily on its data fidelity term and achieves a more accurate colorization. In parallel, this weighting is illustrated by our $\phi$ map and can be beneficial for users or downstream tasks such as computer vision ones.}

\newcommand{\subfigBIGPtwo}[2]{ \subfigure[#1]{ \includegraphics[width=0.27\linewidth, trim={0 0 0 0}, clip]{#2} } }
\newcommand{\subfigBIGPphitwo}[2]{ \subfigure[#1]{ \includegraphics[width=0.27\linewidth, trim={43 43 43 43}, clip]{#2} } }

\begin{figure}[t]
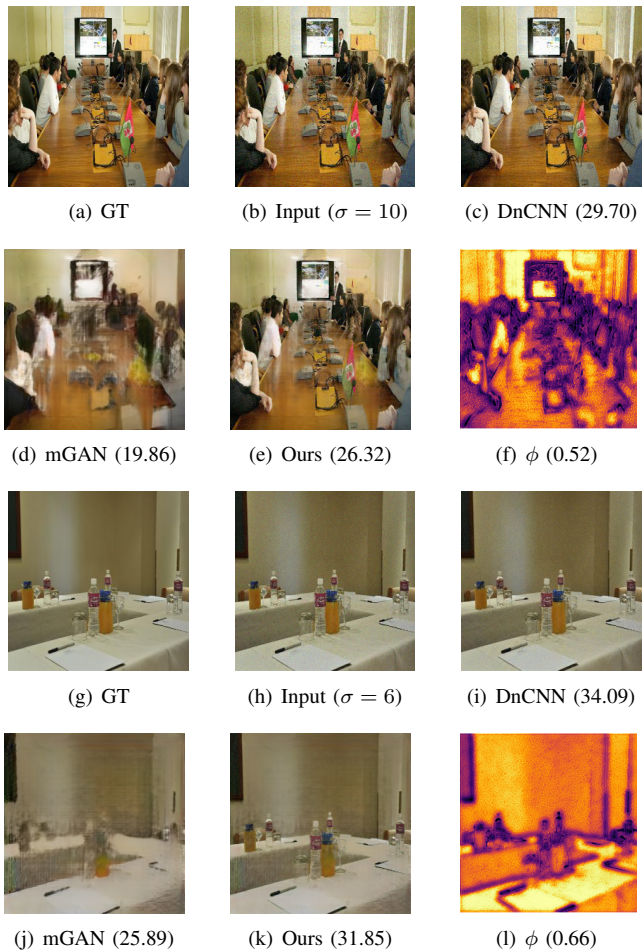

	\centering
	\subfigBIGPtwo{GT}{AWGN/251_gt.png}
	\subfigBIGPtwo{Input ($\sigma=10$)}{AWGN/251_inp.png}
	\subfigBIGPtwo{DnCNN (29.70)}{AWGN/251_dncnn.png}
	\subfigBIGPtwo{mGAN (19.86)}{AWGN/251_mgan.png}
	\subfigBIGPtwo{Ours (26.32)}{AWGN/251.png}
	\subfigBIGPphitwo{$\phi$ (0.52)}{AWGN/251_phi.png}
	
	\subfigBIGPtwo{GT}{AWGN/269_gt.png}
	\subfigBIGPtwo{Input ($\sigma=6$)}{AWGN/269_inp.png}
	\subfigBIGPtwo{DnCNN (34.09)}{AWGN/269_dncnn.png}
	\subfigBIGPtwo{mGAN (25.89)}{AWGN/269_mgan.png}
	\subfigBIGPtwo{Ours (31.85)}{AWGN/269.png}
	\subfigBIGPphitwo{$\phi$ (0.66)}{AWGN/269_phi.png}
	
	\caption[AWGN removal failure cases]{Failure cases of AWGN removal. The quality of the generative-network inversion, which remains a very challenging task, is detrimental to our final results. Although our results significantly improve on the prior by exploiting the input observation by using our fusion weight, they still fall short of the task-specific DnCNN denoiser's results.} \label{fig:failure_cases_awgn}
\end{figure}

\begin{figure}[t]
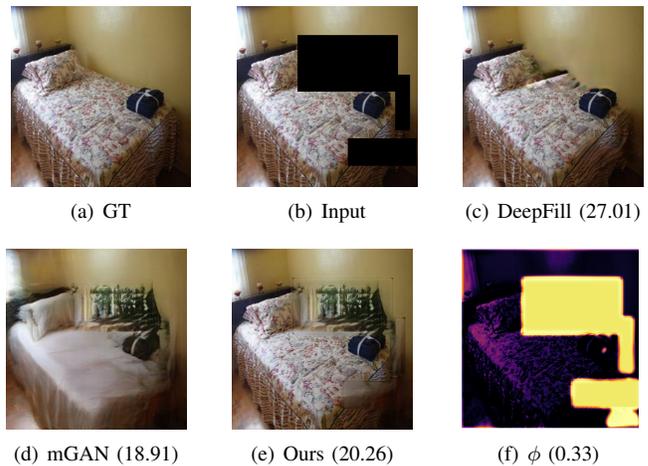

	\centering
	
	\subfigBIGPtwo{GT}{VarInp/293_gt.png}
	\subfigBIGPtwo{Input}{VarInp/293_inp.png}
	\subfigBIGPtwo{DeepFill (27.01)}{VarInp/293_dfill.png}
	\subfigBIGPtwo{mGAN (18.91)}{VarInp/293_mgan.png}
	\subfigBIGPtwo{Ours (20.26)}{VarInp/293.png}
	\subfigBIGPphitwo{$\phi$ (0.33)}{VarInp/293_phi.png} %phi=0.33
	
	\caption[Randomized inpainting failure case]{Failure case in a randomized inpainting experiment. We note a misprediction in the $\phi$ mask in (f), in the bottom right corner. Our network mistakenly assumed the very dark region was a masked region. Note that in inpainting, data fidelity cannot be used in the masked area, and our results become directly dependent on the quality of the prior that, in this example, is not high.} \label{fig:failure_cases_inp}
\end{figure}

\begin{figure*}[t]
	\centering
	\subfigure[Correlation between $\phi$ and the AWGN $\sigma$]{
		\includegraphics[width=.92\linewidth, trim={130 0 135 0}, clip]{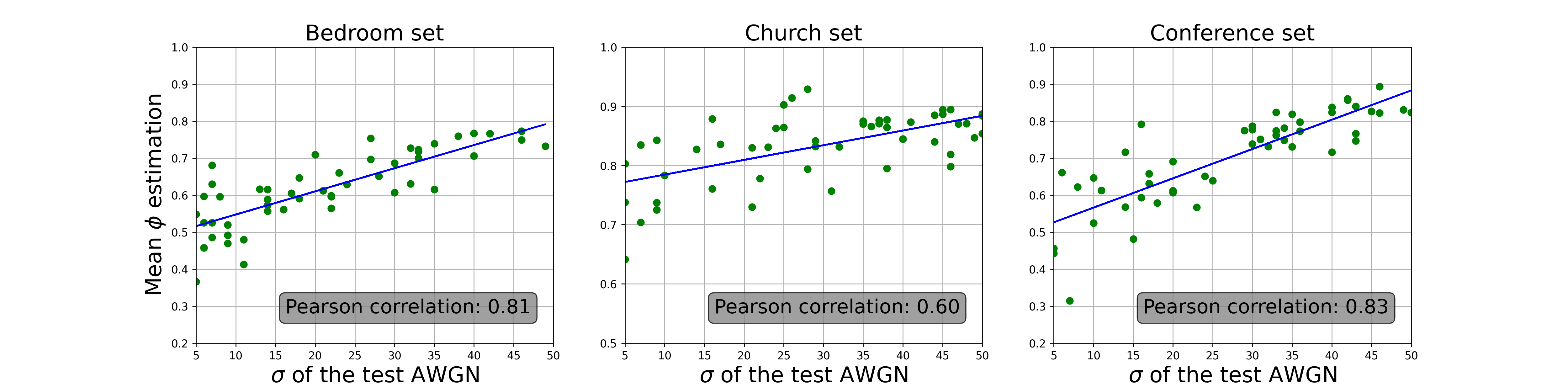}
		\label{fig:phi_corr_sigma}
	}
	\subfigure[Correlation between $\phi$ and the generative PSNR]{
		\includegraphics[width=.92\linewidth, trim={130 0 135 0}, clip]{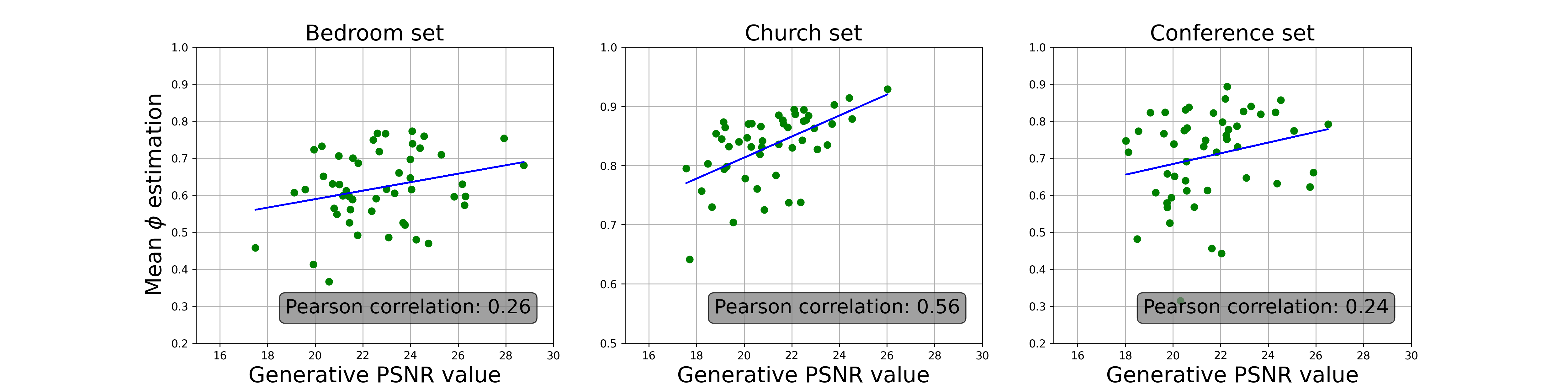}
		\label{fig:phi_corr_psnr}
	}
	\caption[Correlation analysis of data and prior quality to our $\phi$ map]{(a) Shows across three datasets the relation between the AWGN standard deviation in test images, which is directly related to signal quality, and our corresponding mean estimations for $\phi$. (b) Shows the same analysis but with respect to the PSNR of the generative network inversion results, which is directly related to the fitness of the prior. The results show a strong Pearson correlation factor between $\phi$ and signal quality (a), with the remaining factor of variation explained by the fitness of the prior (b) (e.g., Church set).} \label{fig:phi_corr}
\end{figure*}

Visual results are shown in Figure~\ref{fig:inp} for the different methods. For our method, there is little flexibility in terms of the fusion factor $\phi$ for the inpainting tasks, which are tasks with binary degradation, i.e., the signal is either perfectly available or not at all. The $\phi$ map effectively predicts the inpainting mask, a mask that is taken as input in the DeepFill method, and the quality of our results is  tied mostly to those of the generative-network inversion, as can be visually observed.

\subsection{Inpainting}
We present results on the standard central-crop inpainting task in Table~\ref{table:inpcrop_results}. A $64\times 64$ patch is masked from the test image, and the task is to recover the hidden crop. For generative-network inversion, we use the fourth layer of the PGGAN for the feature composition, with 30 latent codes, and $\ell_2$ and VGG-16 perceptual loss~\cite{simonyan2015very}, optimized with gradient descent for 3000 iterations, following~\cite{gu2020image}. We use an identity function $g^{-1}(y)=y$ for the data fidelity, and the remaining setup follows that of Section~\ref{sec:sec:bigp_expsetup}. The PSNR, SSIM and LPIPS results show a significant improvement of our approach, due to the use of the data fidelity, over the mGAN prior results ($+4.77dB$ in terms of PSNR). The inpainting results are averaged across the Bedroom, Church (Outdoor), and Conference datasets. We compare them with the deep-image prior method~\cite{ulyanov2018deep} and with the recent feature map optimization approach~\cite{bau2020semantic} that is a method using GAN priors with test-image specific adaptation. For reference, we compare the results with a task-specific supervised inpainting method, namely, the most recent version~\cite{yu2019free} of DeepFill~\cite{yu2018generative}, trained on the Places2 dataset. DeepFill takes the mask as input and uses gated convolutions to account for invalid pixel locations, and contextual attention~\cite{yu2018generative} to exploit similar patches across the image. The output is refined by using an adversarial GAN loss on every neuron in the feature space~\cite{yu2019free}.
For inpainting, our approach cannot use anything out of the signal over the masked area hence is dependent on the prior hallucination.

The aforementioned benchmarking setup, however, makes the task simpler for our method in terms of predicting $\phi$. Therefore, we design a randomized-masking inpainting setup and present experimental results on it in Table~\ref{table:inpvar_results}. Our randomized-masking algorithm selects uniformly at random a number of patches to be masked, in $[2,4]$. Then, per patch, a random pixel location for the corner of that patch is selected. The algorithm samples from a normal distribution $\mathcal{N}(64,32)$, truncated to $[9,+\infty)$, a width and a height for each patch, with re-sampling in case the patch extends beyond the image coordinates. We compare the mGAN prior results with ours in Table~\ref{table:inpvar_results}. We omit the other methods because the purpose of this randomized-masking experiment is specifically to analyze the effect of randomizing the mask on our $\phi$ prediction, and to analyze how the incurred errors in $\phi$ affect the performance relative to the prior. %DeepFill as its results do not vary significantly since it explicitly takes the test mask as input at inference time. 
We can first note that the mGAN performance decreases, by almost $0.02$ SSIM on average. With the randomization of the mask, our performance decreases even more, by almost $0.1$ SSIM. However, we still significantly improve over the mGAN results, \medits{by  +2.88, +2.62, and +2.82 PSNR on the Bedroom, Church, and Conference test sets, respectively, and +0.270, +0.315, and +0.268 in terms of SSIM on those same datasets}. This comparison highlights the increased difficulty of our internal $\phi$ prediction when the mask is randomized relative to the central inpainting task where the mask location is immutable.

\subsection{Denoising}
\textbf{Blind AWGN Removal.}
We conduct experiments on blind denoising, specifically on AWGN removal. For blind denoising, we follow the standard setup~\cite{dncnn,el2020blind,elhelou2020stochastic} of sampling a noise level, uniformly at random over the range $[5,50]$. This level is the standard deviation of the AWGN. For generative network inversion, we use the fourth layer of the PGGAN for the feature composition, with 30 latent codes, and $\ell_2$ and VGG-16 perceptual loss~\cite{simonyan2015very}, optimized with gradient descent for 3000 iterations, following~\cite{gu2020image}. We set $f(\cdot)$ (Equation~\eqref{eq:general_inversion}) to the identity. Our $g^{-1}(y)$ function is also the identity function as the noise is zero-mean. Generally, $g^{-1}(\cdot)$ can be the subtraction of the noise mean value. For the remaining setup details, we follow the experimental setup of Section~\ref{sec:sec:bigp_expsetup}. 

We present the AWGN removal results in Table~\ref{table:denblind_results}, along with those of DnCNN~\cite{dncnn}, which we retrain on the same data as ours. Our approach achieves the best performance, consistently across the different datasets and evaluation metrics. \medits{For instance, our smallest improvement in terms of PSNR is +2.26$dB$ relative to the mGAN outputs on the Church dataset, and of +0.98$dB$ relative to DnCNN on that same test set. Similarly we also improve the perceptual metrics SSIM and LPIPS on the different test sets.}

Visual results are shown in Figure~\ref{fig:denblind} for the different methods. We observe that DnCNN preserves details well, but at the cost of poorer denoising on low-frequency regions (e.g., walls). The mGAN results are worse than DnCNN, but with our framework the final results become more visually pleasing and accurate. The $\phi$ map illustrates, per pixel, the contribution of hallucination relative to data fidelity and is again lower around edges, as with colorization. We analyze $\phi$ in more detail, in the context of AWGN removal, in the next section.

%# church: mean phi=0.6703
%# conference: mean phi=0.6924
%# bedroom: mean phi=0.7318
\medits{\textbf{Real Denoising on SIDD.} We extend our denoising experiments to the Smartphone Image Denoising Dataset (SIDD) benchmark~\cite{abdelhamed2018high}. We present in Table~\ref{table:sidd} the results of standard classic and learning-based denoising methods, namely, Trainable Nonlinear Reaction Diffusion (TNRD)~\cite{chen2016trainable}, BM3D~\cite{BM3D}, WNNM~\cite{WNNM}, KSVD~\cite{KSVD}, Expected Patch Log Likelihood (EPLL)~\cite{EPLL}, DnCNN~\cite{dncnn}, and CBDNet~\cite{Guo2019Cbdnet}. We also report the results of HI-GAN~\cite{vo2021hi} and the Variational Denoising Network (VDN)~\cite{yue2019variational}. These methods are designed specifically to address the problem of handling real image data, adapting to the distribution of the noise in the SIDD data. For mGAN~\cite{gu2020image} and Ours, we present the results with the different PGGAN generators pre-trained on the Church, Conference, and Bedroom datasets and following the same experimental procedure as for blind AWGN removal. We begin by noting that these two approaches use the small version of the SIDD dataset (SIDD-Small) because of the high computational power required to carry out the generative network inversion and the high resolution of SIDD images. 
Our denoising performance is competitive with standard denoising methods, but it does not reach that of the state-of-the-art task-specific methods HI-GAN and VDN. However, our method additionally provides the $\phi$ map that determines the degree of contribution of prior-based hallucination relative to data fidelity. We can observe that our final PSNR results improve when our method can rely more on its prior (larger $\phi$ in Table~\ref{table:sidd}). This indicates that with better priors, either improved methods or richer pre-training datasets, our method can provide increasingly better denoising results.}

\begin{table}[t!]
	\centering
	\begin{tabular}{lcc}
		\toprule
		& Mean PSNR $\uparrow$ & Mean $\phi$ \\ \toprule % & Mean SSIM $\uparrow$ \\ 
%		\rowcolor{Gray}
		TNRD~\cite{chen2016trainable} & 24.73 & -- \\% & 0.643 \\
		BM3D~\cite{BM3D} & 25.65 & -- \\% & 0.685 \\
		WNNM~\cite{WNNM} & 25.78 & -- \\% & 0.809 \\
		KSVD~\cite{KSVD} & 26.88 & -- \\% & 0.842 \\
		EPLL~\cite{EPLL} & 27.11 & -- \\% & 0.870 \\
		DnCNN~\cite{dncnn} & 32.59 & -- \\% & 0.861 \\
		CBDNet~\cite{Guo2019Cbdnet} & 33.28 & -- \\% & 0.868 \\
		
		\hdashline
		$^\dagger$(Church PGGAN) mGAN~\cite{gu2020image} & 32.66 & -- \\
		(Church PGGAN) Ours & 34.01 & 0.6703 \\
		(Conference PGGAN) mGAN~\cite{gu2020image} & 33.26 & -- \\
		(Conference PGGAN) Ours & 34.30 & 0.6924\\
		(Bedroom PGGAN) mGAN~\cite{gu2020image} & 33.14 & -- \\
		(Bedroom PGGAN) Ours & 34.46 & 0.7318 \\
		\hdashline
		%		FFDNet & 38.27 & 0.948 \\
		HI-GAN~\cite{vo2021hi} & 38.88 & -- \\% & 0.952 \\
		VDN~\cite{yue2019variational} & 39.26 & -- \\% & 0.955 \\
		
		\bottomrule \\
	\end{tabular}
	\caption[SIDD results]{\medits{Quantitative image denoising results on the real image SIDD benchmark~\cite{abdelhamed2018high}. The best results are obtained by HI-GAN~\cite{vo2021hi} and VDN~\cite{yue2019variational} that are specifically designed for real image denoising and capable of learning the noise distribution for each dataset (VDN). $^\dagger$We note that mGAN and Ours use central crops of the small version of the SIDD dataset (SIDD-Small) due to the computational complexity of the underlying generative network inversion, as we discuss in Section~\ref{sec:bigp_discussion}.}}
	\label{table:sidd}
\end{table}

\section{Discussion}~\label{sec:bigp_discussion}
\subsection{Computation Times}
\medits{\textbf{Prior extraction.}
We choose to exploit in our experimental design a generative network inversion for our prior extraction. The mGAN~\cite{gu2020image} inversion on the PGGAN generator takes on average $\approx$0.15sec per inverse iteration on one Titan X GPU. That is $\approx$3.75min for the experiments with 1,500 iterations, and $\approx$7.5min for the 3,000 iteration experiments. 

\textbf{BIGPrior fusion.} Our fusion pipeline with BIGPrior involves efficient operations for applying the bijective $g^{-1}(\cdot)$ function to reverse the degradation, and for applying the $\phi$ map weight to both the data fidelity term and the prior-based term. BIGPrior also requires the prediction of the $\phi$ map, which is carried out using a deep network. The architecture we use is similar to DnCNN~\cite{dncnn} and requires $\approx$0.1sec on a Titan X GPU for inference. 

We can therefore note that although our $\phi$ prediction requires similar computation times as standard feed-forward CNN solutions, the computational bottleneck is due to the network inversion for extracting the prior. Improving the computational efficiency of our presented solution requires more efficient network inversions that are outside the scope of this manuscript, or the use of other prior extraction techniques, for instance, using dictionary-based methods.}

\subsection{Analysis of $\phi$}
The AWGN experiments provide the ideal setup for an analysis of $\phi$ that we carry out in this section. We know that $\phi$ should be inversely related to the signal quality, the poorer the signal is, the higher the $\phi$ values are. And $\phi$ is then also directly related to the confidence in the prior, or the fitness thereof. With AWGN, the quality of the signal is also inversely related to the noise level, in this case, to the standard deviation of the Gaussian noise. We analyze the correlation between the mean $\phi$ value for a test image, and the standard deviation of the noise in this test image. Results are shown in Figure~\ref{fig:phi_corr_sigma}, with the Pearson correlation factor, for three datasets. We can clearly observe the positive correlation between the two variables, with a factor of 0.83 and 0.81 for the bedroom and conference sets, respectively. The correlation is lower, at 0.6, for the church dataset.
The remaining factor of variability in $\phi$ is the fitness of the prior, which we analyze in Figure~\ref{fig:phi_corr_psnr}. The correlation between the average $\phi$ value and the generative PSNR is the highest for the church set, reaching 0.56, and supporting our claims with regard to $\phi$. Indeed, we observe that $\phi$ is well-correlated with the signal's quality, and when that correlation is somewhat lower it is matched with a higher correlation between $\phi$ and the fitness of the prior, exactly as expected from the MAP framework's perspective. To summarize, we make two supporting observations from our aforementioned analysis. First, the $\phi$ estimation, which is learned with no guide in our framework, strongly correlates with the signal quality. Second, a lower correlation with signal quality, as in the church set, is directly justified by a higher correlation between $\phi$ and the fitness of the prior to the test data. These two observations align exactly with the intuitions derived from the MAP estimation framework, as presented in Section~\ref{sec:bigp_method_math}.

The framework we present can be a novel basis for image restoration as it can counter the obstacle of degradation-model overfitting, common in image-restoration tasks. This is because hallucination is the key part prone to overfitting to the chosen model. Our framework can guard against this type of overfitting by relying on decoupled data fidelity and prior hallucination, and by using a pre-trained and frozen generative network, independent of the degradation model, for the hallucination part. Our fusion factor could also be used to increase the robustness and reliability of down-stream computer vision tasks, by making the latter aware of the extent of per-pixel hallucination in the restoration result. For instance, when a computer-vision algorithm deals with degraded images, rather than training the downstream network only on the restoration output, further information regarding the degree of hallucination can be used to increase robustness, notably against adversarial attacks. Our fusion map $\phi$ conveys such hallucination information, which can also be used for better interpretability of the results by human users.

\subsection{Limitations}
\meditstwo{As mentioned earlier in this section, one of the limitations of our current implementation of the BIGPrior framework is the computational overhead required by the prior extraction. However, our framework can be used with any future extraction technique, one option being the learning of the prior projection step to replace it by an efficient feed-forward solution.

Another limitation, also related to the current implementation of the prior projection part, is due to the generative network itself. The training of such networks in an adversarial setup is expensive and requires significant amounts of data to achieve good image distribution learning and in turn good image quality. However, future generative solutions can be integrated into our framework in a straight-forward manner, simply by replacing the pre-trained generator.} %~\cite{karras2021alias}

\section{Conclusion}
We have presented a framework for image restoration that enables the use of deep networks for extracting an image prior while decoupling prior-based hallucination and data fidelity. We have shown how our framework is a generalization of a large family of classic restoration methods, notably of Bayesian MAP estimation setups, and of dictionary-based restoration methods. 

We have conducted experiments on image colorization, inpainting, and Gaussian \medits{and real} denoising. Our results, which structurally come with a pixel-wise map indicating data fidelity versus prior hallucination contributions, outperform prior-based methods and are even competitive with state-of-the-art task-specific supervised methods. We have also presented an analysis of our fusion factor $\phi$ map estimation that supports the claims we make.

% % use section* for acknowledgment
% \section*{Acknowledgment}

{
\bibliographystyle{IEEEtran.bst}
\bibliography{refs.bib}
}

% \begin{IEEEbiographynophoto}{Majed El Helou}
% Biography text here.
% \end{IEEEbiographynophoto}

% %\newpage

% \begin{IEEEbiographynophoto}{Sabine S{\"u}sstrunk}
% Biography text here.
% \end{IEEEbiographynophoto}

\end{document}